\documentclass[11pt,a4paper]{article}
\usepackage[hyperref]{acl2020}
\usepackage{times}
\usepackage{latexsym}

\usepackage{booktabs}
\usepackage{cleveref}
\usepackage{url}
\usepackage[textsize=tiny]{todonotes}

\usepackage{microtype}

\aclfinalcopy 
\def\aclpaperid{1676} 

\newcommand{\mname}{SUPERT}

\title{\mname{}: Towards New Frontiers in Unsupervised 
Evaluation Metrics
for Multi-Document Summarization}

\author{Yang Gao$^{\dagger}$, Wei Zhao$^{\ddagger}$,
Steffen Eger$^{\ddagger}$\\
    $^\dagger $ Dept. of Computer Science, Royal Holloway, University of London, UK \\
    $^\ddagger$ Computer Science Department, Technische Universit\"at Darmstadt, Germany \\
    \texttt{yang.gao@rhul.ac.uk},
     \texttt{\{zhao,eger\}@aiphes.tu-darmstadt.de} \\
  }

\date{}

\begin{document}
\maketitle
\begin{abstract}
We study \emph{unsupervised multi-document summarization evaluation}
metrics, which require neither human-written reference
summaries nor human annotations (e.g.\ preferences, ratings, etc.). 
We propose \emph{\mname{}}, 
which rates the quality of a summary
by measuring its semantic similarity with 
a \emph{pseudo reference summary},
i.e.\ selected salient sentences 
from the source documents,
using contextualized embeddings
and soft token alignment techniques.
%
%
Compared to the state-of-the-art 
unsupervised evaluation metrics, 
\mname{} correlates 
better with human ratings by 18-39\%. 
Furthermore, we use \mname{} as rewards to guide
a neural-based \emph{reinforcement learning} 
summarizer, yielding
favorable performance compared
to the state-of-the-art unsupervised summarizers.
All source code is available at \url{https://github.com/yg211/acl20-ref-free-eval}.
\end{abstract}

\section{Introduction}
\label{sec:intro}

%
Evaluating the quality of machine-generated summaries is a highly laborious
and hence expensive task.  
Most existing evaluation methods require certain forms of
human involvement, thus are \emph{supervised}:
they either directly let humans 
rate the generated summaries (e.g. Pyramid
\cite{DBLP:conf/naacl/NenkovaP04}), 
elicit human-written reference summaries 
and measure their overlap
with the generated summaries
(e.g.\ using ROGUE \cite{lin2004looking}
or MoverScore \cite{zhao-etal-2019-moverscore}), or 
collect some human annotations 
(e.g.\ 
preferences over pairs of summaries \cite{gao-irj-2019})
to learn a summary evaluation function. 
%
Evaluation in \emph{multi-document summarization}
is particularly expensive:
\citet{lin2004rouge} reports that it requires 3,000 hours
of human effort to evaluate the summaries 
from the Document Understanding Conferences
(DUC)\footnote{\url{http://duc.nist.gov/}}.

To reduce the expenses for evaluating multi-document summaries,
we investigate \emph{unsupervised evaluation}
methods, which require neither human annotations nor reference summaries. 
In particular, we focus on evaluating the \emph{relevance}
\cite{peyrard-2019-simple} of 
multi-document summaries, i.e.\ 
measuring how much salient information from the source documents is
covered by the summaries.
%
There exist a few unsupervised evaluation methods
\cite{DBLP:journals/coling/LouisN13,sun-nenkova-2019-feasibility},
but 
they have low correlation with human relevance ratings
at \emph{summary level}:
given multiple summaries for the same source documents, 
these methods can hardly distinguish summaries with
high relevance from 
those with low relevance (see \S\ref{sec:baselines}).

%

\begin{figure}
    \centering
    \includegraphics[width=0.5\textwidth]{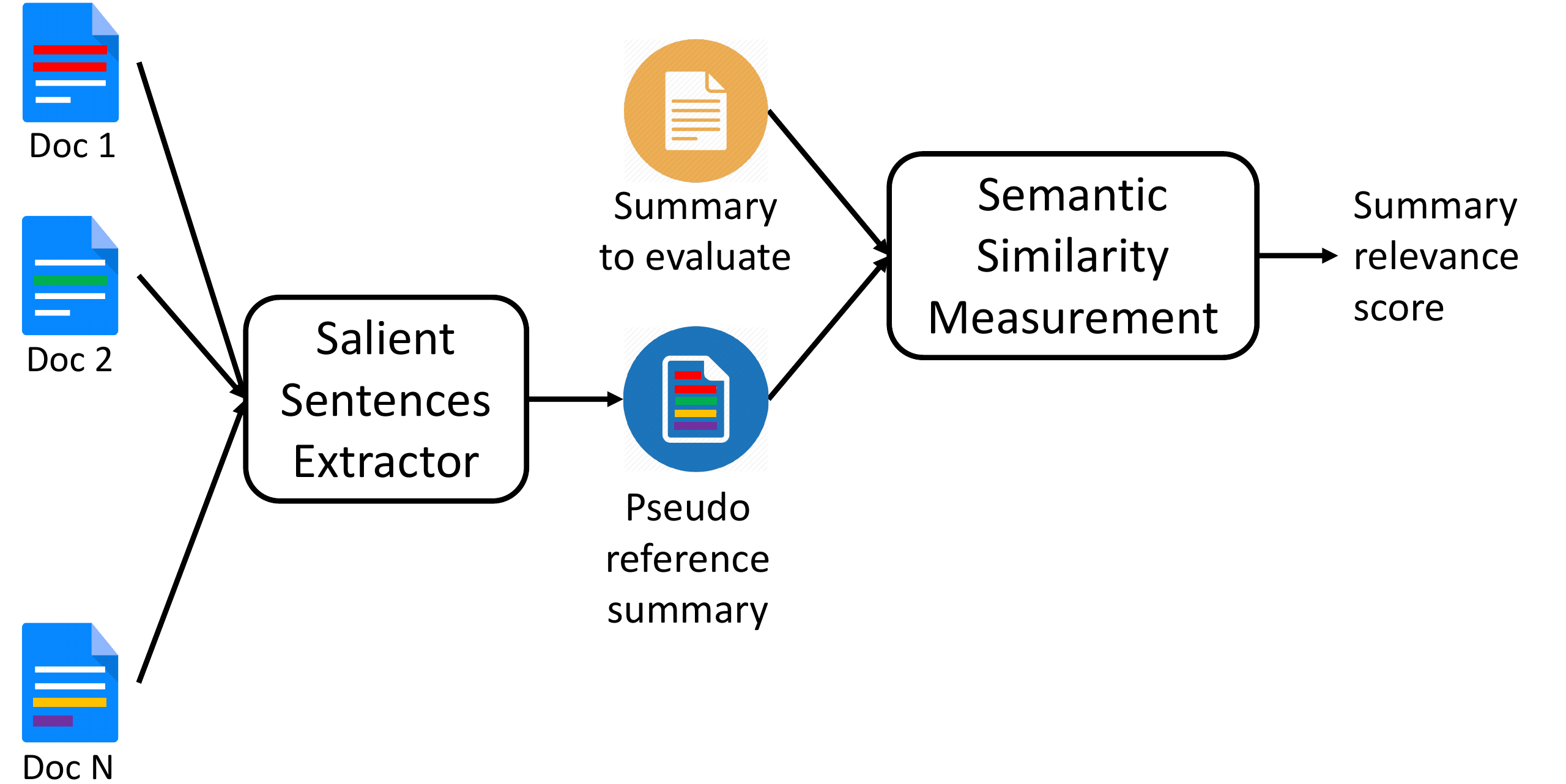}
    \caption{Workflow of \mname{}.}
    \label{fig:workflow}
\end{figure}

\paragraph{Contributions.}
First, to 
better measure the semantic overlap between source documents
and machine-generated summaries,
we propose to use state-of-the-art
contextualized text encoders, e.g.\ 
\emph{BERT} \cite{bert-2019-devlin} 
and its variant \emph{Sentence-BERT} (SBERT) \cite{reimers-gurevych-2019-sentence},
which is optimized for measuring 
semantic similarity between sentences, 
to develop unsupervised evaluation methods. 
We measure the relevance of a summary 
in two steps:
\textbf{(i)} identifying the salient information 
in the input documents, to build a 
\emph{pseudo reference summary},
and \textbf{(ii)} measuring the semantic overlap
between the pseudo reference and the 
summary to be evaluated.
The resulting evaluation method
is called \emph{\mname{}}
(SUmmarization evaluation with Pseudo references and bERT).
Fig.~\ref{fig:workflow} illustrates the 
major steps of \mname{}.
%
We show that compared to 
state-of-the-art unsupervised metrics, 
the best \mname{} correlates 
better 
with the human ratings by 18-39\% 
(in Kendall's $\tau$). 

Second, we use 
\mname{} as 
\emph{reward functions}
to guide \emph{Reinforcement Learning} (RL) 
based extractive summarizers.
%
%
We show it
outperforms the state-of-the-art unsupervised summarization
methods (in multiple ROUGE metrics).
%

\section{Related Work}
\label{sec:related_work}

\paragraph{Reference-based Evaluation.}
Popular metrics like
ROUGE \cite{lin2004looking}, BLEU \cite{DBLP:conf/acl/PapineniRWZ02} 
and METEOR \cite{DBLP:journals/mt/LavieD09} 
fall into this category.
They require (preferably, multiple) 
human written references
and measure the relevance of a summary by 
comparing its overlapping word sequences 
with references.
%
%
More recent work extends ROUGE with WordNet
\cite{shafieibavani-etal-2018-graph}, 
word embeddings \cite{rouge-we-2015}, 
or use 
contextualized-embedding-based methods
\cite{bert-score-2019,zhao-etal-2019-moverscore} to measure
the semantic similarity between references and summaries.
%

\paragraph{Annotation-based Evaluation.} 
%
Some methods directly ask human annotators
to rate summaries following some guidelines, e.g.\
\emph{Responsiveness},
which measures the overall quality (relevance, fluency and readability) of summaries,
and \emph{Pyramid} \cite{DBLP:conf/naacl/NenkovaP04}, which measures  summaries' relevance.
Recently,
systems have been developed to ease 
the construction of Pyramid scores, e.g.\  
\cite{hirao-etal-2018-automatic,DBLP:conf/aaai/YangPM16,gao-etal-2019-automated,ori-2019-lite-pyramid},
but they still require human-annotated 
Summary Content Units (SCUs) to produce
reliable scores.
Besides SCUs, recent work has explored eliciting 
preferences over summaries 
\cite{zopf-2018-estimating,DBLP:conf/emnlp/GaoMG18,gao-irj-2019} and
annotations of important bi-grams \cite{p-v-s-meyer-2017-joint}
to derive summary ratings.

Some methods collect
human ratings on a small number of summaries 
to train an evaluation function. 
\citet{DBLP:conf/emnlp/PeyrardBG17,DBLP:conf/naacl/PeyrardG18}
propose to learn an evaluation function from Pyramid
and Responsiveness scores, by using classic supervised learning 
methods with hand-crafted features.
%
%
\citet{shafieibavani-etal-2018-summarization} use the same idea
but design corpus based and 
lexical resource based word embeddings to build the features.
%
%
\citet{bohm-etal-2019-better} train a BERT-based 
evaluation function with 2,500 human ratings
for 500 machine-generated summaries from 
the CNN/DailyMail dataset;
%
their method correlates better with human ratings
than ROUGE and BLEU.
However, as their method is designed for evaluating
single-document summaries, it correlates poorly 
with the Pyramid scores for multi-document summaries
(see \S\ref{sec:baselines}).
%

\paragraph{Unsupervised Evaluation.}
\citet{DBLP:journals/coling/LouisN13} 
measure the relevance
of a summary using multiple heuristics, for example by 
computing the Jensen-Shannon (JS) 
divergence between the word distributions
in the summary and in the source documents.
\citet{ryang2012emnlp,rioux2014emnlp} develop 
evaluation heuristics inspired 
by the maximal marginal relevance metrics \cite{goldstein-etal-2000-multi}.
But these methods have low correlation with human ratings 
at summary level (see \S\ref{sec:baselines}).
\citet{scialom-etal-2019-answers}
propose to generate questions from
source documents and evaluate the relevance of summaries
by counting how many questions the summaries can answer.
However, they do not detail how to generate  questions
from source documents; also, it remains unclear whether 
their method works for evaluating multi-document summaries.
\citet{sun-nenkova-2019-feasibility} propose
a single-document summary evaluation method, which 
measures the cosine similarity
of the 
ELMo embeddings \cite{peters-etal-2018-deep}
of the source document and the summary.
In \S\ref{sec:baselines}, we show that their 
method performs poorly in evaluating multi-document
summaries. 
\mname{} extends their method by using more advanced
contextualized embeddings and more effective text
alignment/matching methods (\S\ref{sec:bert_baselines}),
and by introducing pseudo references 
(\S\ref{sec:sent_extraction}).
%

\section{Datasets, Baselines and Upper Bounds}
\label{sec:baselines}

\paragraph{Datasets.}
We use two multi-document summarization 
datasets from the 
Text Analysis Conference (TAC)\footnote{\url{https://tac.nist.gov/}} 
shared tasks: TAC'08 and TAC'09. 
In line with \citet{DBLP:journals/coling/LouisN13}, 
we only use the initial summaries (the A part) in these datasets.
TAC'08 includes 48 topics and TAC'09 includes 44. 
Each topic has ten news articles, four reference summaries
and 57 (TAC'08) and 55 (TAC'09) machine-generated summaries.
Each news article on average has 611 words in 24 sentences.
Each summary has at most 100 words and receives 
a Pyramid score, which is used
as the ground-truth human rating in our experiments.
%

\paragraph{Baselines \& Upper Bounds.}
For baselines, we consider 
\textit{TF-IDF}, which computes the cosine similarity
of the tf-idf vectors of source and summaries;
\textit{JS}, which
computes the JS divergence between the words 
distributions in source 
documents and summaries; and 
the \textit{REAPER} heuristics proposed by \citet{rioux2014emnlp}.
In addition, we use the learned metric from
\citet{bohm-etal-2019-better} (\emph{B{\" o}hm19}) and 
the ELMo-based metric 
by \citet{sun-nenkova-2019-feasibility} (\emph{C\textsubscript{ELMo}},
stands for cosine-ELMo; 
see \S\ref{sec:related_work}).
In all these methods, 
we remove stop-words and use the stemmed words,
as we find these operations improve the performance.
For C\textsubscript{ELMo}, we vectorize the documents/summaries
by averaging their sentences' ELMo embeddings.
As for upper bounds, we 
consider three strong 
reference-based evaluation metrics:
ROUGE-1/2 and MoverScore \cite{zhao-etal-2019-moverscore};
note that references are not available
for unsupervised evaluation metrics.

\begin{table}[t]
    \centering
    \small
    \begin{tabular}{l  r r r | r r r}
    \toprule
    & \multicolumn{3}{c|}{TAC'08} & \multicolumn{3}{c}{TAC'09} \\
     & $r$ & $\rho$ & $\tau$ & $r$ & $\rho$ & $\tau$  \\
    \midrule
    \multicolumn{7}{l}{\emph{Baselines (unsupervised evaluation)}} \\
    TF-IDF & .364 & .330 & .236 & \textbf{.388} & \textbf{.395} & \textbf{.288}\\
    JS & \textbf{.381} & \textbf{.333} & \textbf{.238} & \textbf{.388} & .386 & .283 \\
    REAPER & .259 & .247 & .174 & .332 & .354 & .252 \\
    C\textsubscript{ELMo} & .139 & .108 & .076 & .334 & .255 & .183 \\
    B{\"o}hm19 & .022 & -.001 & .001 & .075 & .043 & .031 \\
    \midrule
    \multicolumn{7}{l}{\emph{Upper bounds (reference-based evaluation)}} \\
    Rouge1 & .747 & .632 & .501 & .808 & .692 & .533\\ 
    Rouge2 & .718 & .635 & .498 & .803 & .694 & .531 \\
    Mover & \textbf{.760} & \textbf{.672} & \textbf{.507} & \textbf{.831} & \textbf{.701} & \textbf{.550}\\
    \bottomrule
    \end{tabular}
    \caption{Summary-level correlation between
    some popular evaluation metrics and human
    ratings. Unsupervised metrics (upper)
    measure the similarity between summaries
    and the source documents, while reference-based
    metrics (bottom) measure the 
    similarity between
    summaries and human-written reference summaries.
    }
    \label{table:baselines}
\end{table}

We measure the performance of the baselines and upper bounds
by their average summary-level correlation with Pyramid,
in terms of Pearson's ($r$), Spearman's ($\rho$) 
and Kendall's ($\tau$) correlation 
coefficients.\footnote{We have also considered
the percentage of significantly correlated
topics; results can be found in the
Github repository.}
Table~\ref{table:baselines} presents the 
results. All baseline methods fall far behind the upper bounds.
Among baselines, the embedding-based methods
(B{\"o}hm19 and C\textsubscript{ELMo}) perform
worse than the other lexical-based baselines. 
%
This observation suggests that to 
rate multi-document summaries, 
using existing single-document summaries
evaluation metrics (B{\"o}hm19) or
computing source-summary embeddings' cosine 
similarity (C\textsubscript{ELMo}) is ineffective.
%


\section{Measuring Similarity with Contextualized Embeddings}
\label{sec:bert_baselines}
In this section, we explore the use of more advanced
contextualized embeddings and more sophisticated
embedding alignment/matching methods (rather than cosine
similarity) to measure summaries relevance.
We first extend C\textsubscript{ELMo} by considering more
contextualized text encoders:
BERT, RoBERTa \cite{liu-etal-2019-roberta}, 
ALBERT \cite{albert-2019} and 
\emph{SBERT}\footnote{
Model \texttt{bert-large-nli-stsb-mean-tokens}.}.
%
We use these encoders to produce embeddings for 
each sentence in the documents/summaries, and perform
average pooling to obtain the vector representations 
for the documents/summaries. We measure the
relevance of a summary by computing the cosine
similarity between its 
embedding and the embedding of the source documents.
%
The upper part in
Table~\ref{table:bert_baselines} presents
the results.
C\textsubscript{SBERT} outperforms the other cosine-embedding
based metrics by a large margin, 
but compared to the lexical-based metrics
(see Table~\ref{table:baselines})
its performance still falls short.
%

\begin{table}[t]
    \centering
    \small
    \begin{tabular}{l  r r r | r r r}
    \toprule
    & \multicolumn{3}{c|}{TAC'08} & \multicolumn{3}{c}{TAC'09} \\
    & $r$ & $\rho$ & $\tau$ & $r$ & $\rho$ & $\tau$  \\
    \midrule
    C\textsubscript{BERT} & .035 & .066 & .048 & .130 & .099 & .071 \\
    C\textsubscript{RoBERTa} & .100 & .126 & .091 & .262 & .233 & .165\\
    C\textsubscript{ALBERT} & .152 & .122 & .086 & .247 & .219 & .157\\
    C\textsubscript{SBERT} & .304 & .269 & .191 & .371 & .319 & .229 \\
    \midrule
    M\textsubscript{RoBERTa}& .366 & .326 & .235 & .357 & .316 & .229\\
    M\textsubscript{SBERT} & \textbf{.466} & \textbf{.428} & \textbf{.311} & \textbf{.436} & \textbf{.435} & \textbf{.320} \\
    \bottomrule
    \end{tabular}
    \caption{Performance of contextual-embedding-based
    metrics. Soft aligning the embeddings of the
    source documents and the summaries (the bottom part)
    yields higher correlation than simply
    computing the embeddings cosine similarity 
    (the upper part).
    }
    \label{table:bert_baselines}
\end{table}

\citet{zhao-etal-2019-moverscore} recently show that, to measure
the semantic similarity between two documents, 
instead of computing their document embeddings cosine similarity,
minimizing their token embeddings 
\emph{word mover's distances} (WMDs) 
\citep{DBLP:conf/icml/KusnerSKW15} yields stronger performance.
By minimizing WMDs, tokens from different documents 
are \emph{soft-aligned}, i.e.\ a token from one document can be aligned to 
multiple relevant tokens from the other document.
We adopt the same idea to measure the semantic similarity
between summaries and source documents, using 
RoBERTa and SBERT (denoted by M\textsubscript{RoBERTa}
and M\textsubscript{SBERT}, respectively).
The bottom part in
Table~\ref{table:bert_baselines} presents the 
results.
The WMD-based scores substantially outperform their cosine-embedding
counterparts; in particular, M\textsubscript{SBERT} 
outperforms all lexical-based baselines 
in Table~\ref{table:baselines}. 
This finding suggests that, to rate multi-document summaries,
soft word alignment methods should be used
on top of contextualized embeddings to achieve good performance.

\section{Building Pseudo References}
\label{sec:sent_extraction}
WMD-based metrics yield the highest correlation
in both \emph{reference-based} (bottom
row in Table~\ref{table:baselines}) and 
\emph{reference-free}
(bottom row in Table~\ref{table:bert_baselines})
settings, but there exists a large gap
between their correlation scores.
This observation highlights the need for 
reference summaries. 
In this section, we explore multiple heuristics 
to build \emph{pseudo references}.

%

\subsection{Simple heuristics}
We first consider two simple strategies to 
build pseudo references: randomly extracting $N$ 
sentences or extracting the first $N$ sentences from
each source document.
Results, presented in Table~\ref{table:position}, 
suggest that
extracting the top 10-15 sentences as
the pseudo references yields 
strong performance: it
outperforms the lexical-based baselines 
(upper part in Table~\ref{table:baselines}) by over 16\%
and M\textsubscript{SBERT} (Table~\ref{table:bert_baselines})
by over 4\%.
These findings confirm the \emph{position bias} in 
news articles (c.f.\ 
\cite{jung-etal-2019-earlier}).

\begin{table}[]
    \centering
    \small
    \begin{tabular}{l  r r r | r r r}
    \toprule
    & \multicolumn{3}{c|}{TAC'08} & \multicolumn{3}{c}{TAC'09} \\
     & $r$ & $\rho$ & $\tau$ & $r$ & $\rho$ & $\tau$ \\
    \midrule
    Random3 & .139 & .194 & .189 & .123 & .172 & .175 \\
    Random5 & .144 & .203 & .199 & .147 & .204 & .206 \\
    Random10 & .163 & .228 & .229 & .201 & .279 & .284 \\
    Random15 & .206 & .287 & .320 & .185 & .258 & .268 \\
    \midrule 
    Top3 & .449 & .408 & .295 & .378 & .390 & .291 \\
    Top5 & .477 & .437 & .316 & .413 & .421 & .314 \\
    Top10 & \textbf{.492} & \textbf{.455} & \textbf{.332} & .444 & .450 & .333\\
    Top15 & .489 & .450 & .327 & \textbf{.454} & \textbf{.459} & \textbf{.340} \\
    \bottomrule
    \end{tabular}
    \caption{Building pseudo references by extracting randomly selected sentences (upper) or the first few sentences (bottom). Results of the random extraction methods
    are averaged over ten independent runs.}
    \label{table:position}
\end{table}


\subsection{Graph-based heuristics} 
Graph-based methods have long been used 
to select salient information from documents,
e.g. \cite{DBLP:journals/jair/ErkanR04,zheng-lapata-2019-sentence}.
These methods build grahs to represent
the source documents, in which
each vertex represents
a sentence and the weight of each edge 
is decided by
the similarity of the corresponding sentence pair. 
Below, we explore two families of graph-based methods
to build pseudo references:
\emph{position-agnostic} and 
\emph{position-aware} graphs, which ignore and consider
the sentences' positional information, respectively.
%

\paragraph{Position-Agnostic Graphs.} 
The first graph we consider is 
SBERT-based LexRank (\emph{SLR}),
which extends the classic LexRank \cite{DBLP:journals/jair/ErkanR04} method 
by measuring the similarity of sentences using
SBERT embeddings cosine similarity.
In addition, we propose an SBERT-based 
clustering (\emph{SC}) method to build graphs,
which first measures the similarity
of sentence pairs 
using SBERT,
and then clusters sentences by using the 
\emph{affinity propagation} \cite{frey2007clustering} 
clustering algorithm;
the center of each cluster is selected to
build the pseudo reference.
We choose affinity propagation 
because it does not require a preset cluster
number (unlike K-Means) and
it automatically finds the center point of each cluster.


For each method (SLR or SC), we consider two variants: 
the \emph{individual-graph} version, 
which builds a graph for each source document
and selects top-$K$ sentences (SLR) or
the centers (SC) from each graph;
and the \emph{global-graph} version, which builds
a graph considering all sentences across all source 
documents for the same topic, and selects 
the top-$M$ sentences (SLR)
or all the centers (SC) in this large graph.
According to our preliminary experiments on 20 randomly
sampled topics, we set $K=10$ and $M=90$. 

\paragraph{Position-Aware Graphs.}
PacSum is a recently proposed graph-based 
method to select salient sentences
from multiple documents 
\cite{zheng-lapata-2019-sentence}.
In PacSum,
a sentence is more likely to be selected
if it has higher average similarity with its
succeeding sentences and lower average similarity with
its preceding sentences.
This strategy allows PacSum to prioritize
the selection of early-position
and ``semantically central'' sentences.
%
We further extend PacSum by using SBERT to 
measure sentences similarity 
(the resulting method is denoted as \emph{SPS})
and consider both the individual- and 
global-graph versions of SPS.

Furthermore, we propose a method called 
\emph{Top+Clique} (\emph{TC}),
which selects the top-$N$ sentences and
the semantically central non-top-$N$ sentences
to build the pseudo references.
TC adopts the following steps:
\textbf{(i)} 
Label top-$N$ sentences from each document as salient.
\textbf{(ii)} 
With the remaining (non-top-$N$) sentences,
build a graph such that only ``highly similar'' 
sentences have an edge between them.
\textbf{(iii)} 
Obtain the cliques from the graph 
and select the semantically central sentence 
(i.e.\
the sentence with highest average similarity
with other sentences in the clique) from
each clique as 
\emph{potentially salient sentences}.
\textbf{(iv)} 
For each potentially salient sentence, 
label it as salient 
if it is not highly similar to any top-$N$
sentences.
%
%
%
Based on preliminary experiments on 20 topics,
we let $N=10$ and the threshold value be $0.75$
for ``highly similar''.

%
%

\begin{table}[]
    \centering
    \small
    \begin{tabular}{l  r r r | r r r}
    \toprule
    & \multicolumn{3}{c|}{TAC'08} & \multicolumn{3}{c}{TAC'09} \\
     & $r$ & $\rho$ & $\tau$ & $r$ & $\rho$ & $\tau$ \\
    \midrule
    \multicolumn{7}{l}{\emph{Position-agnostic graphs}} \\
    SLR\textsubscript{I} & .456  & .417 & .304 & .415 & \textbf{.423} & \textbf{.311}  \\
    SLR\textsubscript{G} & \textbf{.461} & \textbf{.423} & \textbf{.306} & \textbf{.419} & \textbf{.423} & .310 \\
    SC\textsubscript{I} & .409 & .364 & .261 & .393 & .383 & .280 \\
    SC\textsubscript{G} & .383 & .344 & .245 & .373 & .365 & .265 \\
    \midrule
    \multicolumn{7}{l}{\emph{Position-aware graphs}} \\
    SPS\textsubscript{I} & .478 & .437 & .319 & .429 & .435 & .321 \\ 
    SPS\textsubscript{G} & .472 & .432 & .313 & .427 & .432 & .318 \\
    TC & \textbf{.490} & \textbf{.449} & \textbf{.329} 
    & \textbf{.450} & \textbf{.454} & \textbf{.336}\\
    \bottomrule
    \end{tabular}
    \caption{
    Building pseudo references by position-agnostic
    (upper) and position-aware (bottom) graphs.
    }
    \label{table:graph}
\end{table}

Table \ref{table:graph} presents the graph-based methods' performance. 
Except for SC\textsubscript{G}, all other graph-based methods 
outperform baselines in Table \ref{table:baselines}.
Position-agnostic graph-based methods perform worse 
not only than the the position-aware ones, 
but even than
the best method in Table~\ref{table:bert_baselines},
which simply uses the full source documents as pseudo references. 
In addition, we find that the position-aware graph-based sentence
extraction methods perform worse than simply extracting top sentences
(Table~\ref{table:position}).
These findings indicate that the position bias remains the most effective
heuristic in selecting salient information from news articles;
when position information is unavailable
(e.g. sentences in source documents are randomly shuffled),
it might be better to use all sentences rather than selecting a subset of 
sentences from the source to build pseudo references.

\section{Guiding Reinforcement Learning}
\label{sec:rl}

We explore the use of different rewards to guide
\emph{Neural Temporal Difference} (NTD),
a  RL-based multi-document summarizer \cite{gao-irj-2019}.
We consider three unsupervised reward functions:
two baseline methods REAPER and JS (see
\S\ref{sec:baselines} and Table~\ref{table:baselines}),
and the best version of \mname{},
which selects the top 10 (TAC'08) 
or 15 (TAC'09) sentences
from each source document to build 
pseudo references and uses 
SBERT to measure the similarity between
summaries and pseudo references.

In addition, we consider a non-RL-based 
state-of-the-art
unsupervised summarizer proposed by \citet{yogatama-etal-2015-extractive} 
(\emph{YLS15}).
%
We use ROUGE to measure
the quality of the generated summaries and leave human 
evaluations for future work.
Table \ref{table:rl_results} presents the results.
We find \mname{} is the strongest reward
among the considered rewards:
it helps NTD perform on par with YSL15
on TAC'08 and perform significantly better on TAC'09.

%
%

\begin{table}[]
    \centering
    \small
    \begin{tabular}{l l l l | l l l}
    \toprule
    & \multicolumn{3}{c|}{TAC'08} & \multicolumn{3}{c}{TAC'09} \\
     & R\textsubscript{1} & R\textsubscript{2} & R\textsubscript{L} &  R\textsubscript{1} & R\textsubscript{2} & R\textsubscript{L} \\
    \midrule
    NTD\textsubscript{RP} & .348 & .087 & .276 & .360 & .090 & .187 \\
    NTD\textsubscript{JS} & .353 & .090 & .281 & .368 & .095 & .192 \\
    NTD\textsubscript{SP} & \textbf{.376}$^*$ & \textbf{.102}$^*$ &  \textbf{.296}$^*$ & \textbf{.380}$^*$ & \textbf{.103}$^*$ & \textbf{.194} \\
    YLS15 & .375$^*$ & .096 & N/A  & .344 & .088 & N/A \\
    \bottomrule
    \end{tabular}
    \caption{Training NTD, a RL-based summarizer, with
    different rewards (RP: REAPER, SP: \mname{}).
    NTD performance is averaged over ten runs. 
    R\textsubscript{1/2/L} stands for ROUGE-1/2/L. 
    $^*$: significant advantage ($p\!\!<\!\!0.01$ double-tailed t-tests)
    over the non-asterisks.}
    \label{table:rl_results}
\end{table}

\section{Conclusion}
\label{sec:conclusion}
We explored unsupervised multi-document summary
evaluation methods, which require neither reference summaries
nor human annotations. 
%
We find that vectorizing the summary and the top sentences in the source
documents using contextualized embeddings,
and measuring their semantic overlap with soft token alignment techniques
is a simple yet effective method to 
rate the summary's quality.
The resulting method,
\emph{\mname{}},
correlates with human ratings 
substantially better than the state-of-the-art unsupervised metrics.

Furthermore, we use \mname{} as rewards to train
a neural-RL-based summarizer, which leads to
up to 17\% quality improvement
(in ROUGE-2) compared to the
state-of-the-art unsupervised summarizers.
This result not only shows the effectiveness of 
\mname{} in a downstream task,
but also promises a new way to 
train RL-based summarizers: 
an infinite number of summary-reward pairs 
can be created from infintely many documents,
and their \mname{} scores can be used
as rewards to train RL-based summarizers,
fundamentally relieving the 
\emph{data-hungriness} 
problem faced by existing RL-based summarization
systems.

\bibliographystyle{acl_natbib}
\bibliography{ref}

\end{document}